%% file: main.tex
\documentclass[twocolumn, 10pt]{article}
\usepackage{graphicx} 

\usepackage{cite}
\usepackage{amsmath,amssymb,amsfonts}
\usepackage{algorithmic}
\usepackage{graphicx}
\usepackage{textcomp}
\usepackage{xcolor}
\usepackage{float}
\usepackage{comment}
\usepackage{enumitem}
\usepackage{amssymb}
\usepackage{mathrsfs}
\usepackage{amsmath}
\usepackage{geometry}
\usepackage{setspace}
\usepackage{enumitem}
\usepackage{amssymb}
\usepackage{setspace}
\usepackage{adjustbox}
\usepackage{float}
\usepackage{booktabs}
\usepackage{amsmath}
\usepackage{xcolor}
\usepackage{acronym}
\usepackage{url}
\usepackage{relsize}
\usepackage{tabularx}

\usepackage[pdfborder={0 0 0}, colorlinks=true, linkcolor=black, citecolor=black, urlcolor= black]{hyperref}
\def\BibTeX{{\rm B\kern-.05em{\sc i\kern-.025em b}\kern-.08em
    T\kern-.1667em\lower.7ex\hbox{E}\kern-.125emX}}

\makeatletter
\renewenvironment{abstract}{%
  \par\small
  \noindent\textbf{Abstract—}\ignorespaces
}{%
  \par\normalsize\bfseries
}
\makeatother

\usepackage{tikz}

\newcommand{\copyrightnotice}{
\begin{tikzpicture}[remember picture,overlay]
\node[anchor=south,yshift=10pt] at (current page.south) {
\fbox{\parbox{\dimexpr\textwidth-2\fboxsep-2\fboxrule\relax}{
\footnotesize
© 2024 IEEE. Personal use of this material is permitted. Permission from IEEE must be obtained for all other uses, in any current or future media, including reprinting/republishing this material for advertising or promotional purposes, creating new collective works, for resale or redistribution to servers or lists, or reuse of any copyrighted component of this work in other works.

This is the Author Accepted Manuscript version of a paper accepted for publication in the \href{https://doi.org/10.1109/NILES63360.2024}{ IEEE 6th Novel Intelligent and Leading Emerging Sciences Conference (NILES)}, 2024. The published version is available via DOI: \href{https://doi.org/10.1109/NILES63360.2024.10753187}{10.1109/NILES63360.2024.10753187}.
}}
};
\end{tikzpicture}
}

\usepackage{titlesec}

\renewcommand{\thesection}{\Roman{section}}
\renewcommand{\thesubsection}{\Alph{subsection}}
\renewcommand{\thesubsubsection}{\arabic{subsubsection}}

\titleformat{\section}
  {\normalfont\filcenter\MakeUppercase}
  {\thesection.}
  {0.6em}
  {}

\titleformat{\subsection}
  {\normalfont\itshape}
  {\thesubsection.}
  {0.6em}
  {}

\titleformat{\subsubsection}[runin]
  {\normalfont\itshape}
  {\thesubsubsection)}
  {0.6em}
  {}
  [.\ ] 

\titlespacing*{\section}{0pt}{1.2ex plus 0.3ex minus 0.2ex}{0.8ex}
\titlespacing*{\subsection}{0pt}{1.0ex plus 0.3ex minus 0.2ex}{0.6ex}
\titlespacing*{\subsubsection}{0pt}{0.8ex plus 0.2ex minus 0.2ex}{0.6em}

\begin{document}

\input{Title}
\copyrightnotice
\input{Abstract}

\input{Introduction}

\input{Methodology}

\input{Results}

\input{Conclusion}

\bibliographystyle{ieeetr}
\bibliography{references}
\end{document}

%% file: Title.tex
\makeatletter
\renewcommand{\maketitle}{%
  \twocolumn[{%
    \begin{@twocolumnfalse}
      \vspace*{-1.2em}
      \begin{center}
        {\huge\bfseries \@title \par} 
        \vspace{0.8em}
        {\normalsize
        \begin{tabular}{@{}ccc@{}}
          \begin{tabular}{@{}c@{}}
            \textbf{Youssef Mahran}\\
            \textit{Mechatronics Engineering Department}\\
            \textit{The German University in Cairo}\\
            Cairo, Egypt\\
            \texttt{youssef.mahran@student.guc.edu.eg}
          \end{tabular}
          &
          \begin{tabular}{@{}c@{}}
            \textbf{Zeyad Gamal}\\
            \textit{Mechatronics Engineering Department}\\
            \textit{The German University in Cairo}\\
            Cairo, Egypt\\
            \texttt{zeyad.abdrabo@student.guc.edu.eg}
          \end{tabular}
          &
          \begin{tabular}{@{}c@{}}
            \textbf{Ayman El-Badawy}\\
            \textit{Mechatronics Engineering Department}\\
            \textit{The German University in Cairo}\\
            Cairo, Egypt\\
            \texttt{ayman.elbadawy@guc.edu.eg}
          \end{tabular}
        \end{tabular}\par
        }
      \end{center}
      \vspace{0.8em}
    \end{@twocolumnfalse}
  }]%
}
\makeatother

\title{Reinforcement Learning Position Control of a Quadrotor Using Soft Actor-Critic (SAC)}
\date{} 

\maketitle

%% file: Abstract.tex
\begin{abstract}
\bfseries This paper proposes a new Reinforcement Learning (RL) based control architecture for quadrotors. With the literature focusing on controlling the four rotors' RPMs directly, this paper aims to control the quadrotor's thrust vector. The RL agent computes the percentage of overall thrust along the quadrotor's $z$-axis along with the desired Roll ($\phi$) and Pitch ($\theta$) angles. The agent then sends the calculated control signals along with the current quadrotor's Yaw angle ($\psi$) to an attitude PID controller. The PID controller then maps the control signals to motor RPMs. The Soft Actor-Critic algorithm, a model-free off-policy stochastic RL algorithm, was used to train the RL agents. Training results show the faster training time of the proposed thrust vector controller in comparison to the conventional RPM controllers. Simulation results show smoother and more accurate path-following for the proposed thrust vector controller.
\end{abstract}

%% file: Introduction.tex
    \section{Introduction}

Quadrotors have become essential nowadays in many applications and became an integral part of many industries. However due to their unstable nature, advanced control algorithms are needed to successfully control the quadrotors. Among various control frameworks, Reinforcement learning (RL) has emerged as a powerful tool. RL controllers allow quadrotors to learn and improve their flying capabilities autonomously using feedback from their environment. Thus, allowing quadrotors to carry out complex tasks with high precision.

Using the Soft Actor-Critic (SAC) algorithm, a low-level controller for a quadrotor that maps environment states to motor commands was proposed \cite{DBLP:journals/corr/abs-2010-02293}. The simulation results prove the controller's effectiveness in an easy go-to task and in tracking moving objects as efficiently as possible at both high and low speeds. The robustness of the agent was then evaluated by testing it with a range of extreme initial positions. In order to maintain control over a quadrotor even in the event of a single rotor failure, an additional SAC agent was created \cite{DBLP:journals/corr/abs-2109-10488}. The SAC's ability to hover, land, and track different trajectories with just three active rotors was shown through the simulation results. To prove the controller's robustness, it was also successfully tested against wind disturbances. A modified SAC algorithm was used to control a quadrotor in precision approach mission and the moving target-chasing mission \cite{drones7090549}. The Step-wise Soft Actor-Critic technique (SeSAC) performs stepwise learning, starting from easier missions, to overcome the inefficiency of learning caused by attempting challenging tasks from the beginning. The experimental results demonstrate that the proposed algorithm successfully completed missions in two challenging scenarios, one for disaster management and another for counter-terrorism missions, while surpassing the performance of other baseline approaches.

This paper aims to develop a cascaded control architecture for low-level control of quadrotors. In this architecture, the process begins with the RL agent calculating the overall thrust percentage, as well as the desired roll and pitch angles. Then, these calculated values are sent to an attitude Proportional-Derivative-Integral Gain (PID) controller, alongside the current yaw angle. The PID controller uses these signals to generate the RPMs for the four rotors. This proposed cascaded approach was never proposed in the literature and it aims to integrate the advantages of both RL and PID to enhance the performance and efficiency of quadrotor flight control.

%% file: Methodology.tex
\section{Methodology}
\subsection{Reinforcement Learning Control Framework}
\begin{figure}[H]
\centerline{\includegraphics[width=0.4\textwidth]{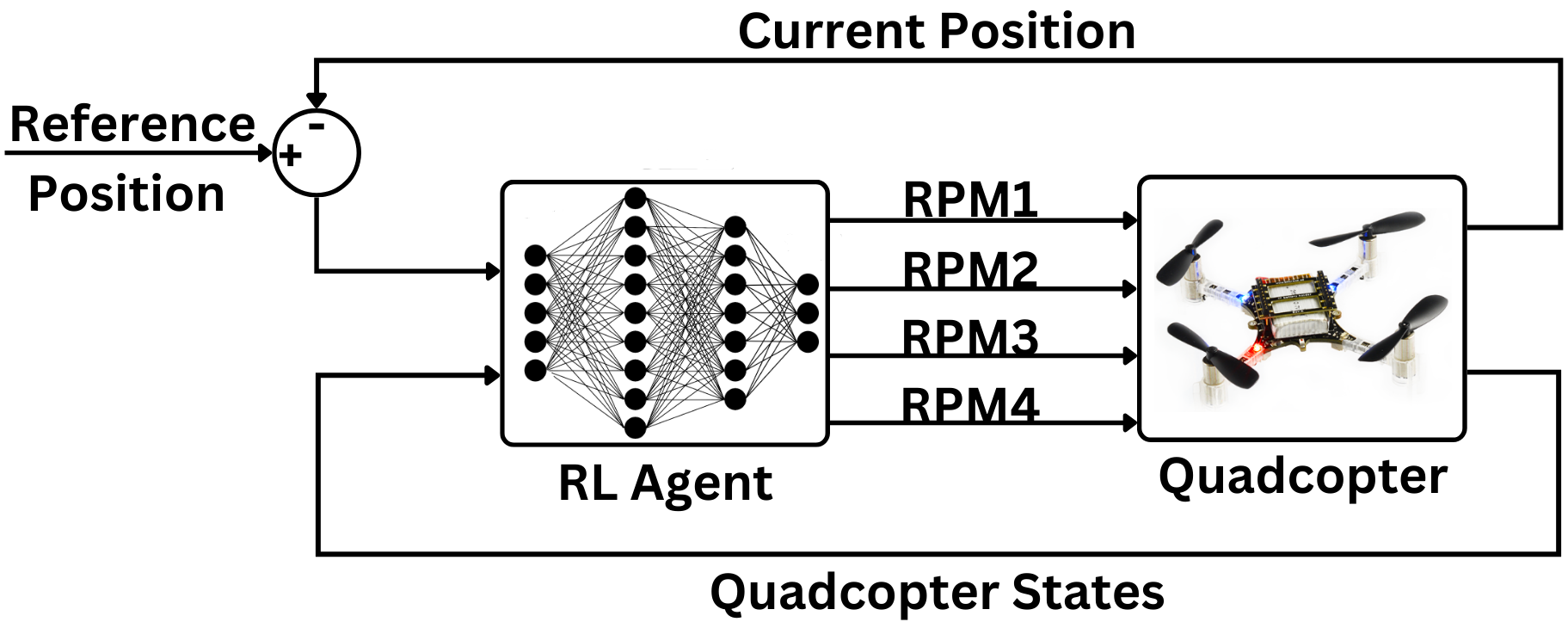}}
\caption{Block diagram of low-level RL RPM Controller}
\label{mokhtar}
\end{figure}

The control framework used in the literature is an RL framework that controls the RPM of the rotors directly \cite{mokhtar2023autonomous}. The RL agent takes as an input the input state vector ($S$) of the quadrotor shown in Eq. \ref{state} and outputs four motor commands to the quadrotor as illustrated in Fig. \ref{mokhtar}.

The input state vector ($S$) is represented as the current orientation of the quadrotor ($\phi, \theta, \psi$), the current linear velocity ($v_{x}, v_{y}, v_{z}$), the current angular velocity ($\omega_{x}, \omega_{y}, \omega_{z}$) and the position error between the current and desired position ($\Delta x, \Delta y, \Delta z$) as shown in Eq. \ref{state}.
\begin{equation}
        S = [\phi, \theta, \psi, v_{\textit{x}}, v_{\textit{y}}, v_{\textit{z}}, \omega_{\textit{x}}, \omega_{\textit{y}}, \omega_{\textit{z}},\Delta x, \Delta y, \Delta z]
        \label{state}
    \end{equation}

\begin{figure}[H]
\centerline{\includegraphics[width=0.5\textwidth]{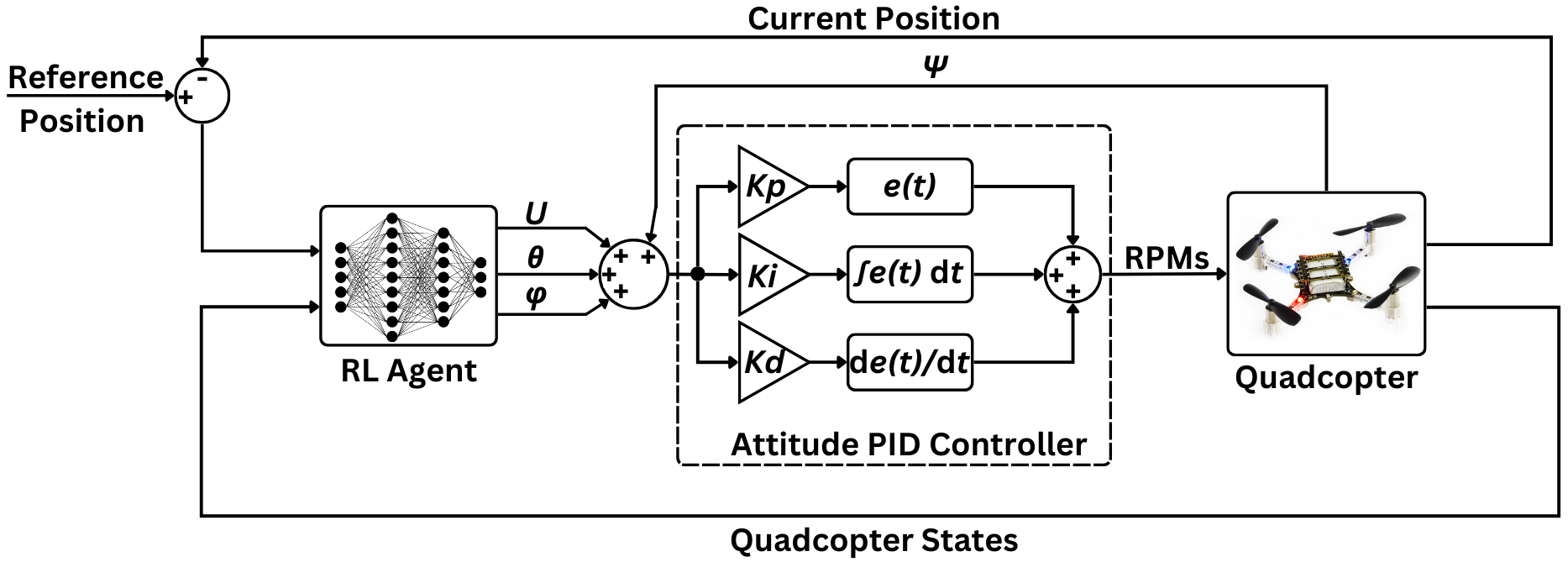}}
\caption{Block diagram of low-level RL Thrust Vector Controller}
\label{mahran}
\end{figure}

In this paper, a new RL control framework is proposed as shown in Fig. \ref{mahran}. In this framework, the RL agent takes the same state vector ($S$) as an input however, the agent computes the percentage of the desired thrust force along the quadrotor's $z$-axis ($U$) and the desired Roll ($\phi$) and Pitch ($\theta$) angles. The calculated actions are sent along with the current Yaw ($\psi$) angle to an attitude PID controller which then maps the signals to motor RPMs.
The two  control frameworks are used in this paper to compare the performance and training time of each controller. The RL algorithm used for both controllers is the SAC algorithm. The SAC is a model-free off-policy stochastic algorithm with dynamic entropy tuning. This algorithm framework allows agents to learn the optimal policy in an efficient manner.

\subsection{SAC Algorithm Network Structure}
  The selected neural network architecture for all networks of the two controllers consists of two hidden layers with 400 and 300 nodes each, one input layer. LeakyReLU activation functions are used in both hidden layers while a $tanh$ activation function is used in the output layer.
 
\subsection{Reward Function}
In reinforcement learning, there are two different kinds of reward functions: dense rewards and sparse rewards. Only in terminal states can the agent obtain sparse rewards. Sparse rewards have significant values that affect the agent's perception of state values as they are usually linked to rare states. On the other hand, dense rewards are often of lesser value and are provided to the agent at each step. Dense reward functions were chosen for the agent since the environment states in the proposed environment are continuous. The reward function was derived from the literature \cite{mokhtar2023autonomous}.
    \begin{equation}
    r = \frac{1}{a \text{  }* \text{  }||\vec{e}_{k}||_{2}} + \frac{a}{\sqrt{2\pi \sigma^2}} e^{-0.5(\frac{||\vec{e}_{k}||_{2}}{\sigma})^2}
    \label{reward}
    \end{equation}
    \begin{equation}
    ||\vec{e}_{k}||_{2} = \sqrt{\Delta x^2 + \Delta y^2 + \Delta z^2}
    \label{ec}
    \end{equation}
    
    There are two different components to the reward function shown in Eq. \ref{reward}. The first component is a rational function, as shown in Eq. \ref{ec}, where $a = 7$ and $||\vec{e}_{k}||_{2}$ denote the Euclidean distance. This component is in charge of monitoring the quadcopter's steady hover. The second component, which tracks the quadcopter, is a normal distribution function with a 0.5 standard deviation. As it gets closer to the target, it receives a larger positive reward \cite{mokhtar2023autonomous}. For both controllers, the same reward function was used.

\subsection{SAC Algorithm Hyperparameters}
A small learning rate was chosen to ensure stability. A high discount factor encourages the agent to favor future rewards over immediate ones. A frequency of 50 Hz was chosen as a suitable frequency for integration with any micro-controller. The same set of hyperparameters is used across the two controllers. The full set of hyperparameters  are shown in Table \ref{tab:ddpg_hyperparameters}.
    \begin{table}[H]
    \centering
    \caption{Hyperparameters used for the SAC algorithm}
    \begin{tabular}{lll}
    \hline
    Symbol & Description & Value \\
    \hline
    $\lambda$ & Learning rate & $0.0007$ \\
    $B$ & Size of the replay buffer &  $1,000,000$\\
    $-$& Steps before learning starts  & $10,000$ \\
    $N$ & Minibatch size & $256$ \\
    $\tau$ & Update coefficient  & $0.005$ \\
    $\gamma$ & Discount factor & $0.99$ \\
    $f$ & Agent frequency  & $50$ Hz \\
    $t_{max}$  & Maximum steps of one episode  & $502$ \\
    \hline
    \end{tabular}
    \label{tab:ddpg_hyperparameters}
    \end{table}

%% file: Results.tex
\section{Results}
\subsection{Stabilization Training}
\begin{figure}[H]
            \centering
            \includegraphics[width=0.5\textwidth]{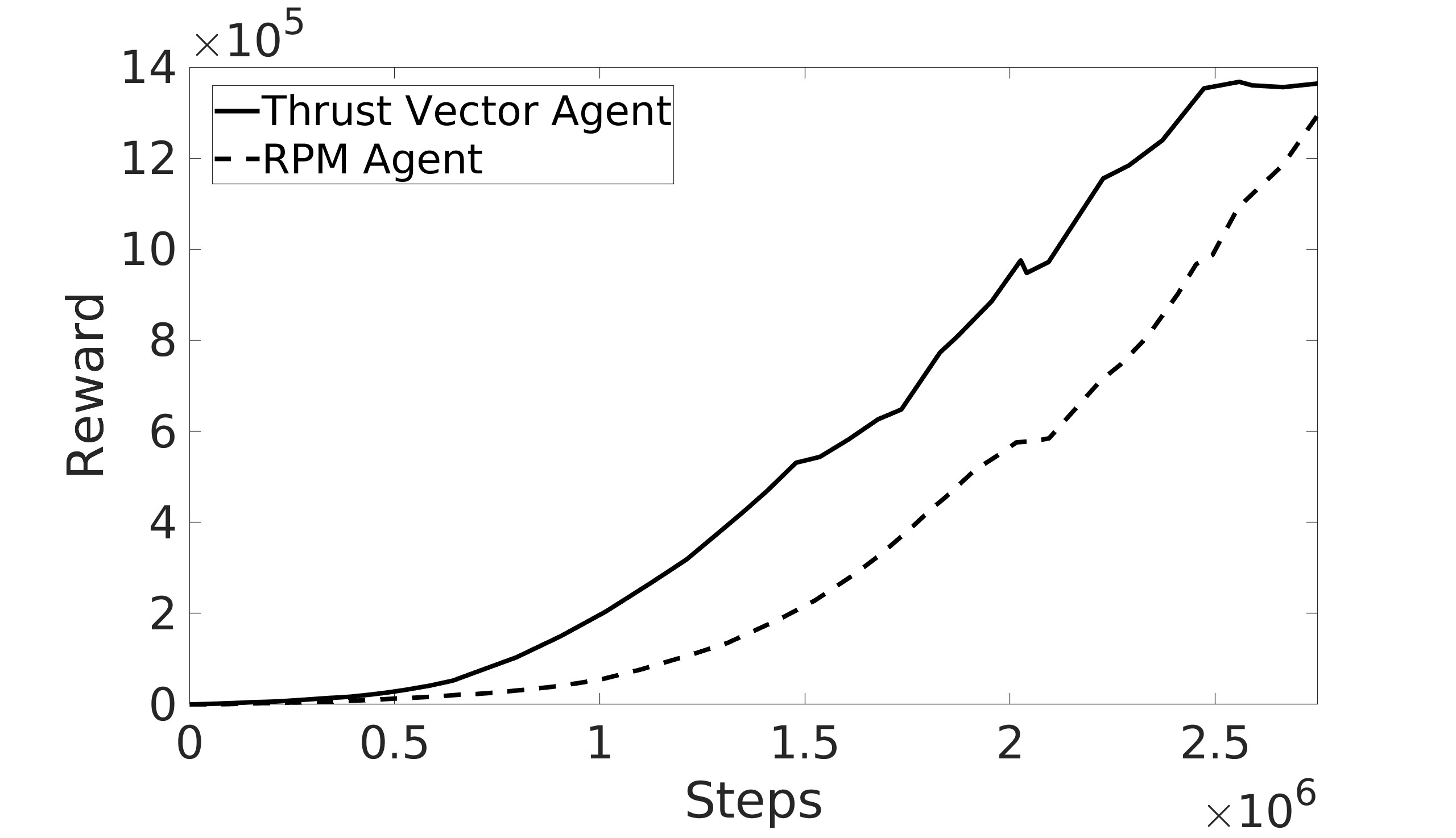}
            \caption{Mean reward for the stabilization training of thrust vector and RPM controllers}
            \label{nonoiserew}
    \end{figure}

The objective of this training was to stabilize the quadrotor at a fixed target of [0, 0, 1] starting from a random position. Both controller agent were trained for 2,750,000 steps. Fig \ref{nonoiserew} shows the mean reward obtained by both agents. Both the proposed thrust vector and RPM agents reached around the same maximum reward at the end of the training.
    
\begin{figure}[H]
            \centering
            \includegraphics[width=0.5\textwidth]{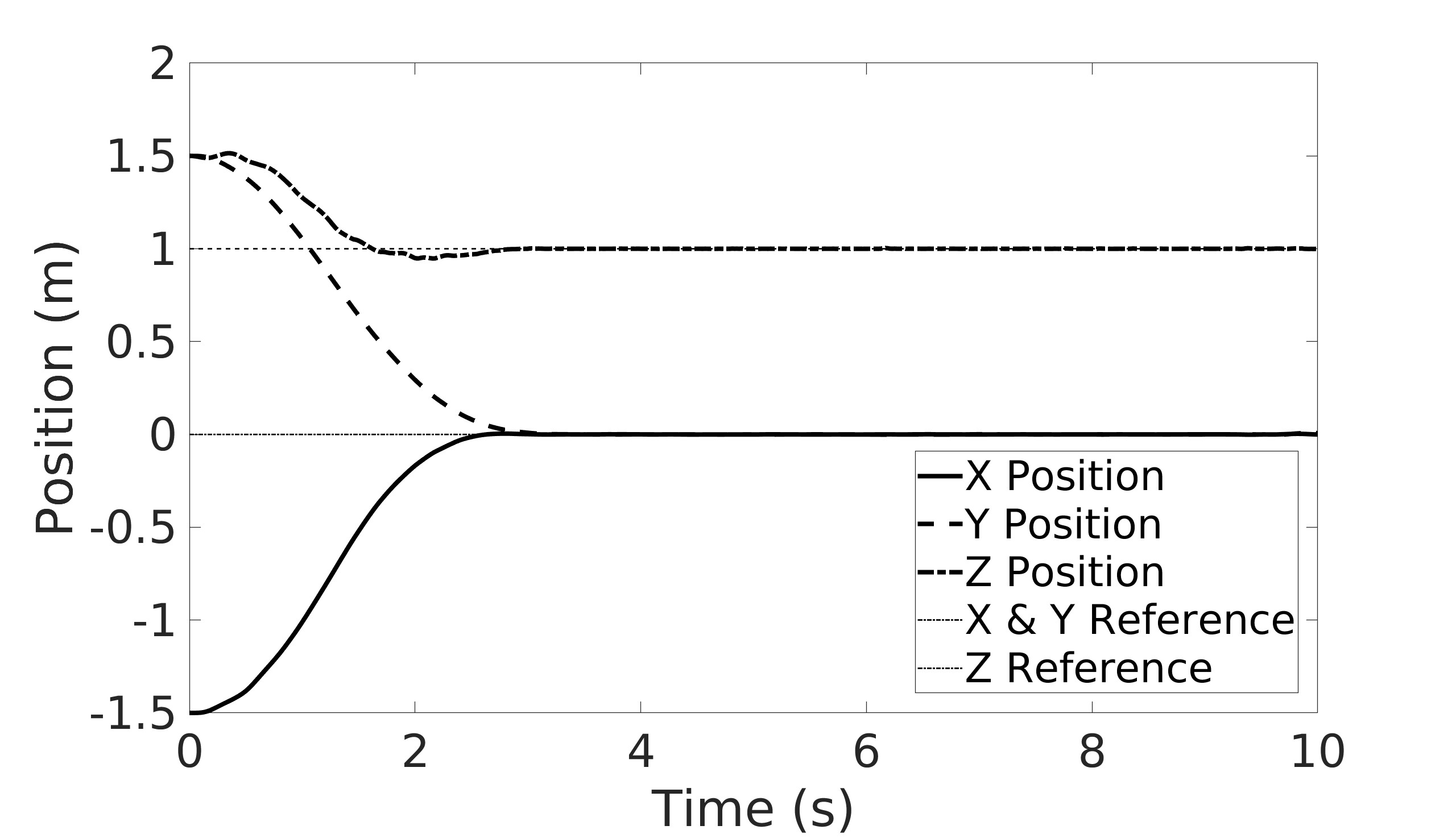}
            \caption{Stabilization response for an initial position of [-1.5, 1.5, 1.5] for thrust vector controller}
            \label{stabu}
    \end{figure}

    Starting from an initial position of [-1.5, 1.5, 1.5], Fig. \ref{stabu} shows the thrust vector controller response. The proposed controller stabilized in all three axes with zero steady-state error.
    
\begin{figure}[H]
            \centering
            \includegraphics[width=0.5\textwidth]{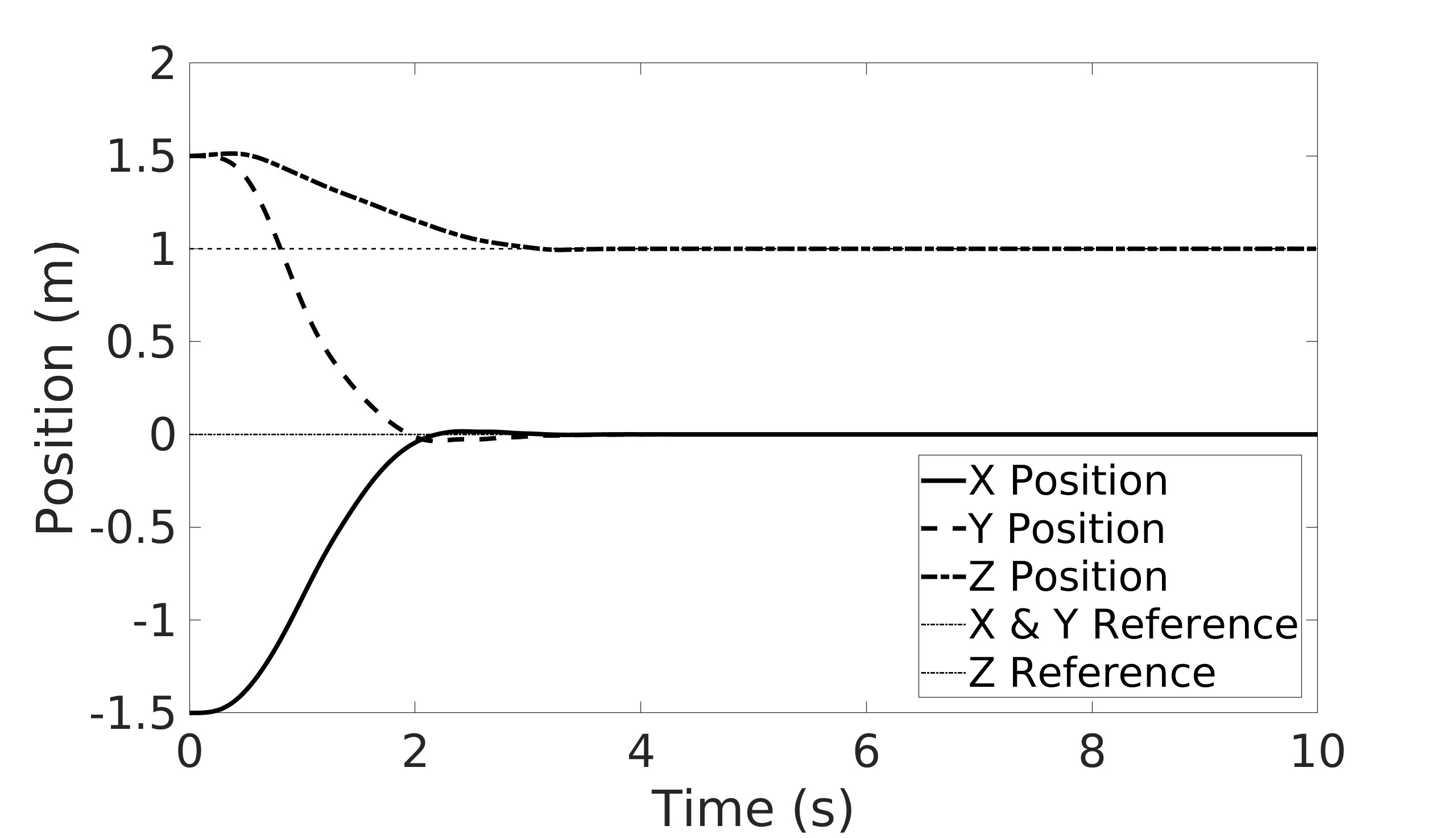}
            \caption{Stabilization response for an initial position of [-1.5, 1.5, 1.5] for RPM controller}
            \label{stabrpm}
    \end{figure}

    Fig. \ref{stabrpm} shows the response of the RPM controller when starting from the same initial position. The controller also stabilized with zero steady-error across the three axes.
    \subsection{Position Tracking Training}
\begin{figure}[H]
            \centering
            \includegraphics[width=0.5\textwidth]{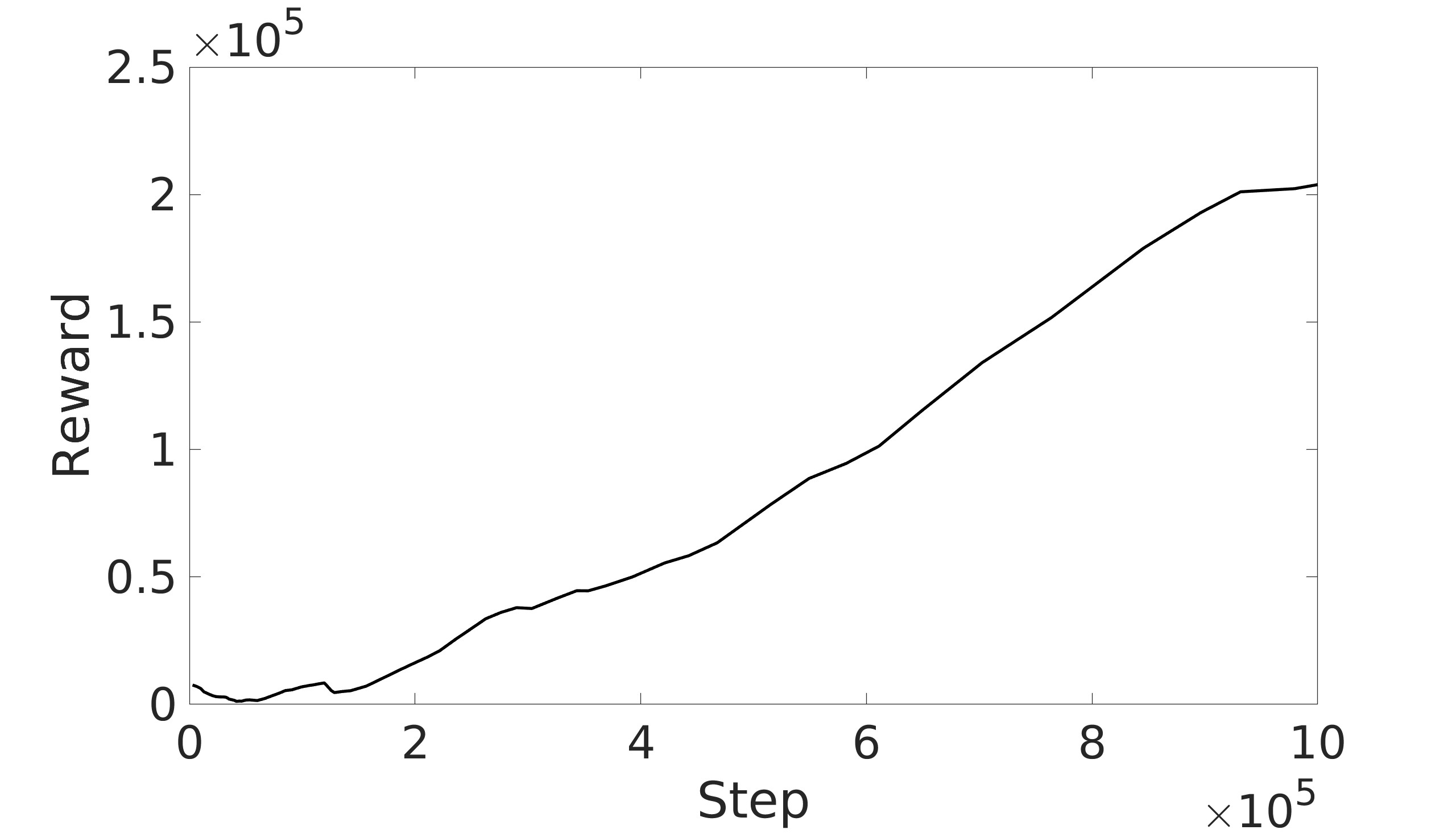}
            \caption{Mean reward for the position tracking training of thrust vector controller}
            \label{trrew}
    \end{figure}

\begin{figure}[H]
            \centering
            \includegraphics[width=0.5\textwidth]{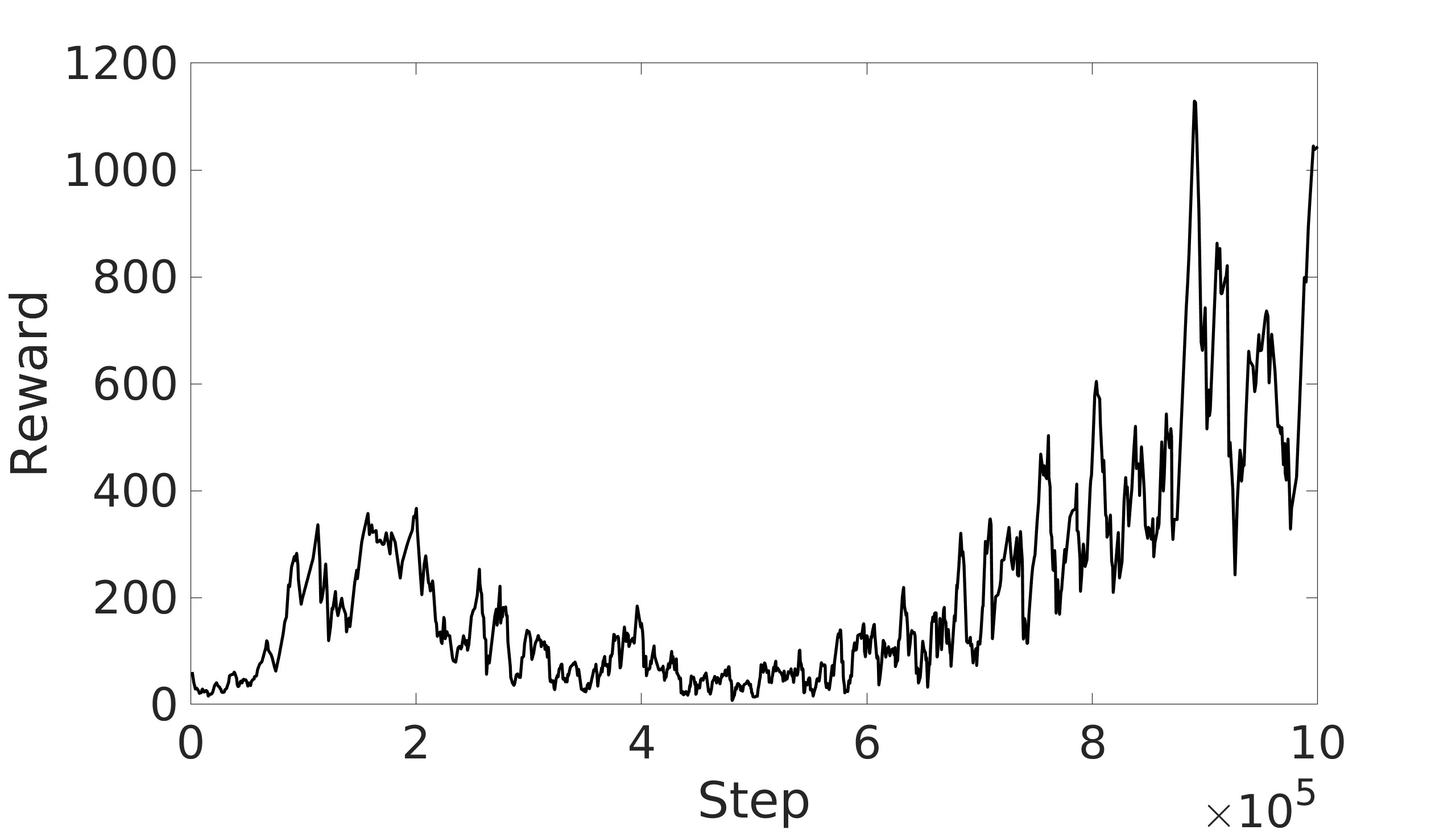}
            \caption{Mean reward for the position tracking training of rpm controller}
            \label{trew}
    \end{figure}

\begin{figure}[H]
            \centering
            \includegraphics[width=0.5\textwidth]{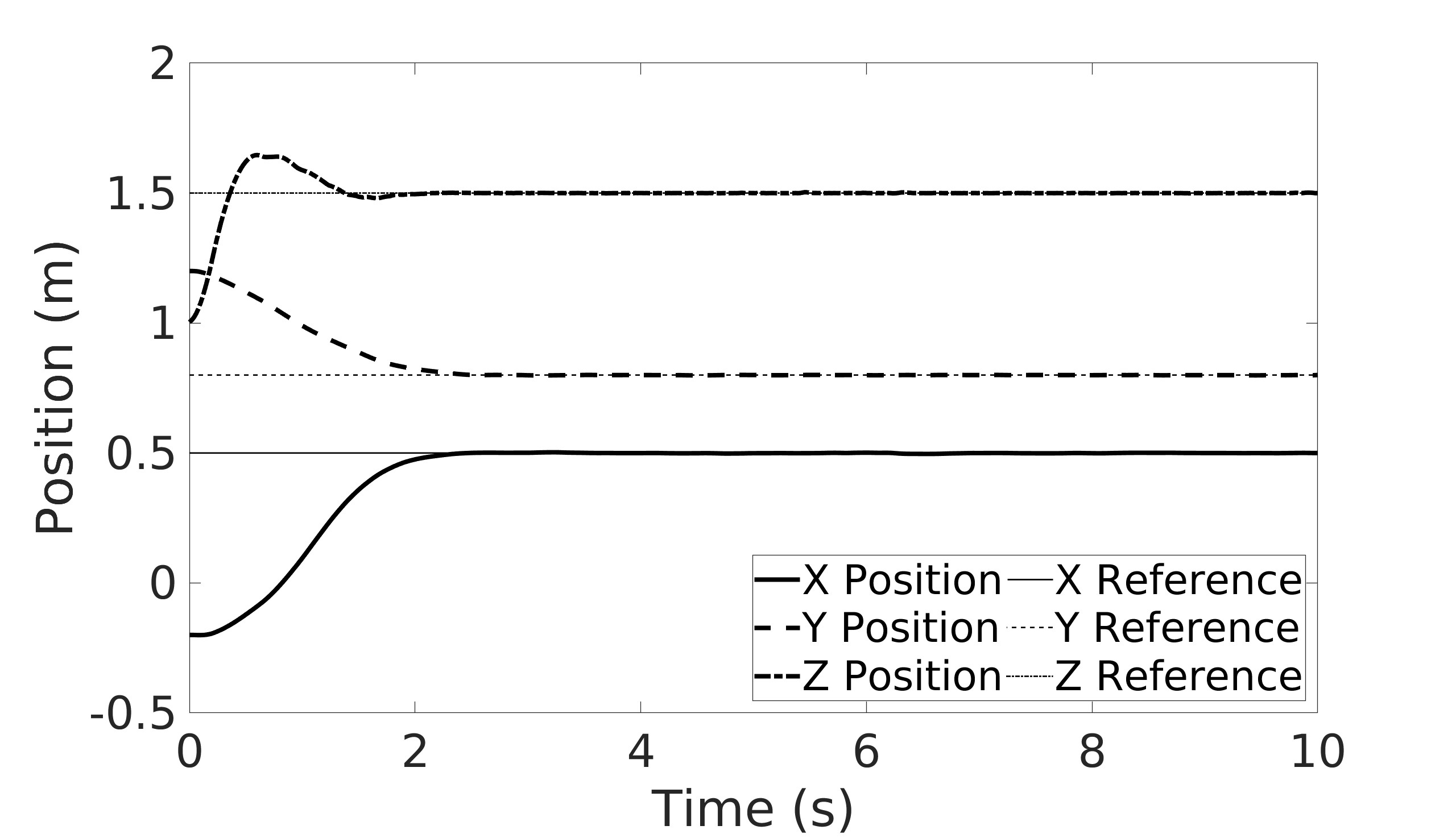}
            \caption{Thrust vector agent response for a target position of [0.5, 0.8, 1.5] and an initial position of [-0.2, 1.2, 1]}
            \label{trackpos1}
    \end{figure}

\begin{figure}[H]
            \centering
            \includegraphics[width=0.5\textwidth]{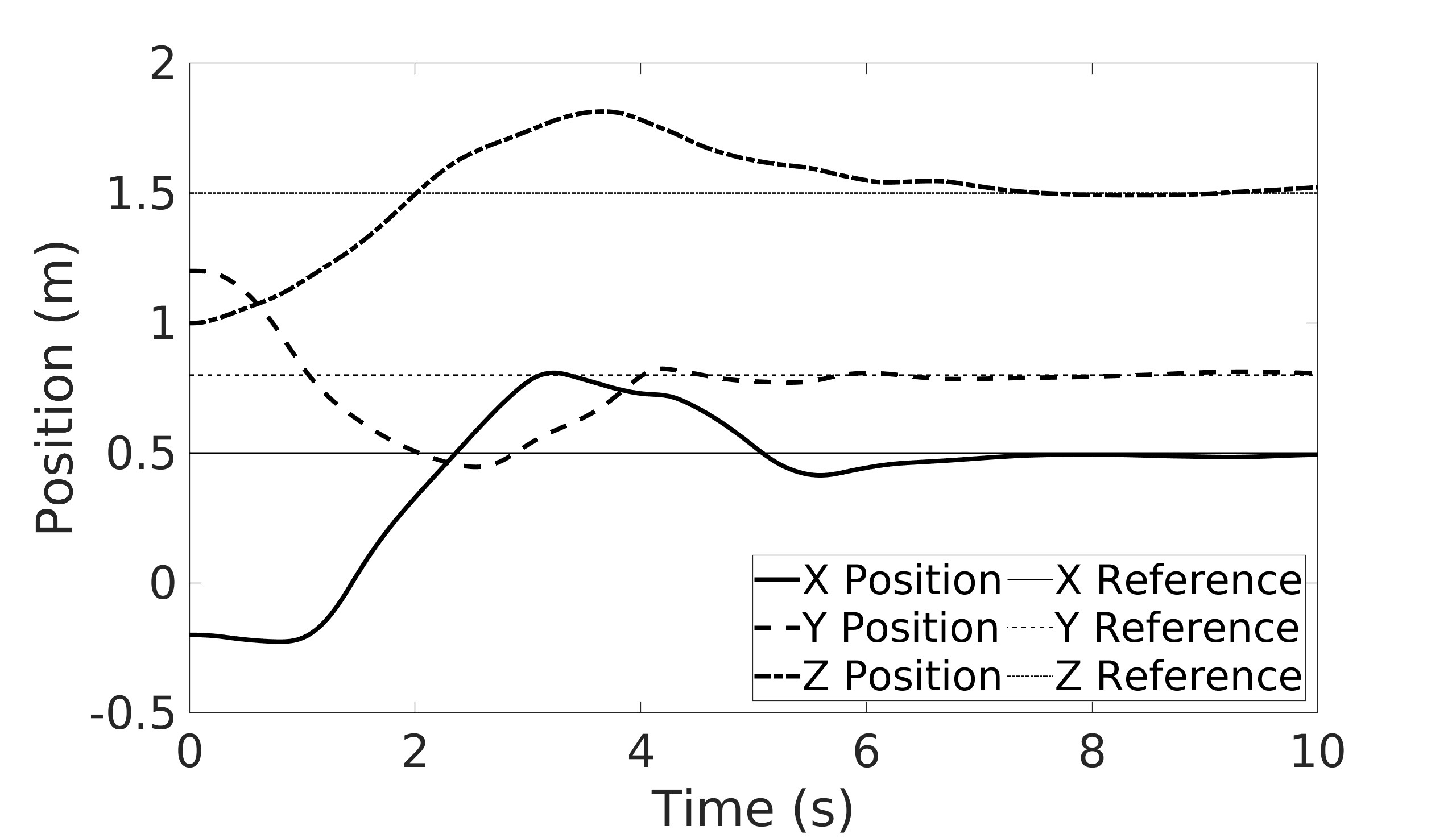}
            \caption{RPM agent response for a target position of [0.5, 0.8, 1.5] and an initial position of [-0.2, 1.2, 1]}
            \label{trpos2}
    \end{figure}

    The objective of this training was for the agent to reach and stabilize the quadrotor starting from a random position however, this time reaching a random target each episode. Fig. \ref{trrew} shows the mean reward obtained by the thrust vector agent. The agent was trained for a total of 1 million steps with the maximum reward of around 20,000 as shown. The thrust vector controller reaches high rewards faster as the agents output is limited to [-1, 1] radians. This helps the agent crash less and stay alive more in the environment to learn better policies.
    Fig. \ref{trew} shows the mean reward obtained by the RPM agent. The agent was trained for the same number of steps with the maximum reward of around 1,100 as shown. The reward obtained by the RPM controller is much lower than the proposed thrust vector controller. Since the agent controls the RPMs directly, it is more vulnerable to crashing and terminating the episode early halting the learning process. This is proven in the literature as the RPM controller requires a longer training time to be able to achieve high rewards as shown in Table \ref{comp}.
     Fig. \ref{trackpos1} shows the thrust vector agent response when starting from an initial position of [-0.2, 1.2, 1] and reaching a target at [0.5, 0.8, 1.5]. The agent stabilized with zero steady-state error in all three axes. The agent had zero overshoot across all axes with a settling time of 2 seconds.
    Fig. \ref{trpos2} shows the RPM agent response when starting from the same initial position and reaching the same target. High overshoots across the three axes occurred. The agent stabilized after 6 seconds with small deviations away from the reference at the end.
    
    \renewcommand{\arraystretch}{2}
    \begin{table*}[t]
    \centering
    \caption{Comparison Between Different Controllers}
    \begin{tabularx}{\textwidth}{XXXXXX}
    \hline
    \textbf{Parameters} & \textbf{Proposed Controller} & \textbf{Shehab Et al. \cite{mazen}} & \textbf{Mokhtar Et al. \cite{mokhtar2023autonomous}} &\textbf{ Barros Et al. \cite{DBLP:journals/corr/abs-2010-02293}} \\
    \hline
    Training Steps & 3.5 million & 15.05 million & 5 million & 3 million\\
    $\substack{\mathlarger{\text{Steady-State}} \\ \mathlarger{\text{Error (X)}}}$ & 0 $m$ &  0.0164 $m$ & 0.025 $m$ & 0.07 $m$\\
    $\substack{\mathlarger{\text{Steady-State}} \\ \mathlarger{\text{Error (Y)}}}$ & 0 $m$ &  0.0308 $m$ & 0.01 $m$ & 0.01 $m$\\
    $\substack{\mathlarger{\text{Steady-State}} \\ \mathlarger{\text{Error (Z)}}}$ & 0 $m$ &  0.0652 $m$ & 0.01 $m$ & 0.2 $m$\\
    \hline
    \end{tabularx}
    \label{comp}
    \end{table*}
    \subsection{Path-Following}
\begin{figure}[H]
            \centering
            \includegraphics[width=0.45\textwidth]{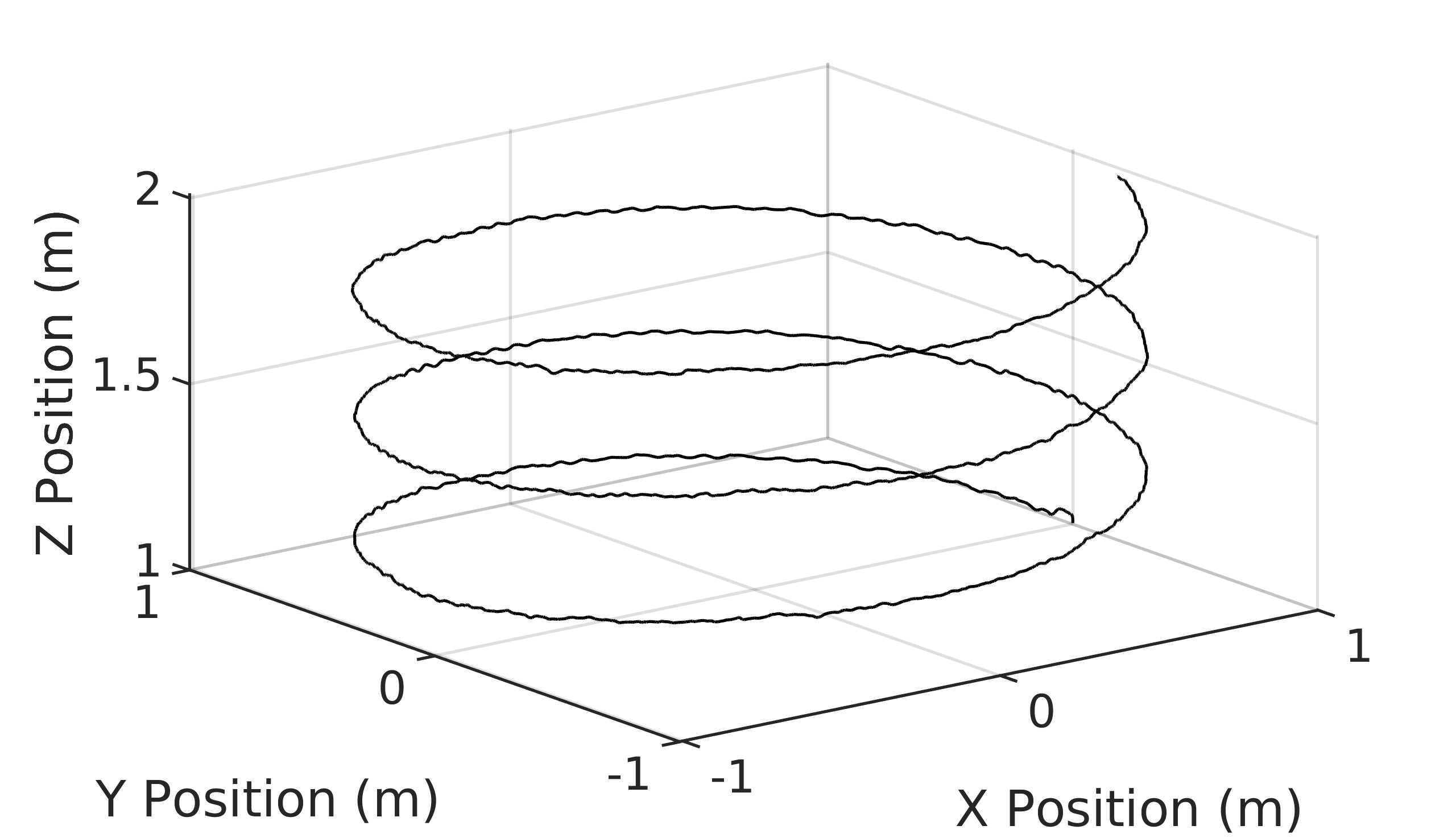}
            \caption{Helix path-following of thrust vector controller}
            \label{traj2}
    \end{figure}

\begin{figure}[H]
            \centering
            \includegraphics[width=0.45\textwidth]{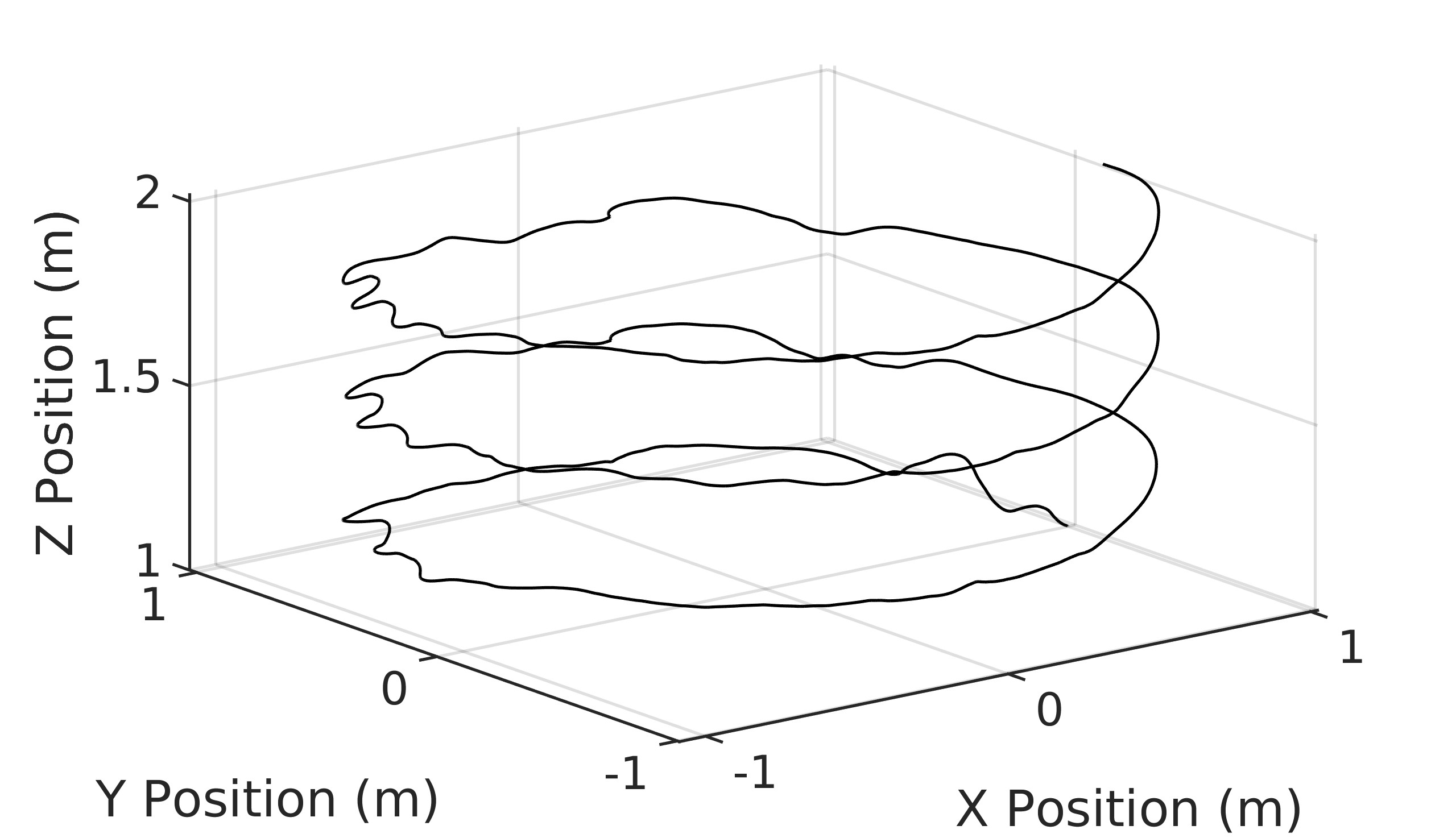}
            \caption{Helix path-following of RPM controller}
            \label{mazinf}
    \end{figure}

\begin{figure}[H]
            \centering
            \includegraphics[width=0.45\textwidth]{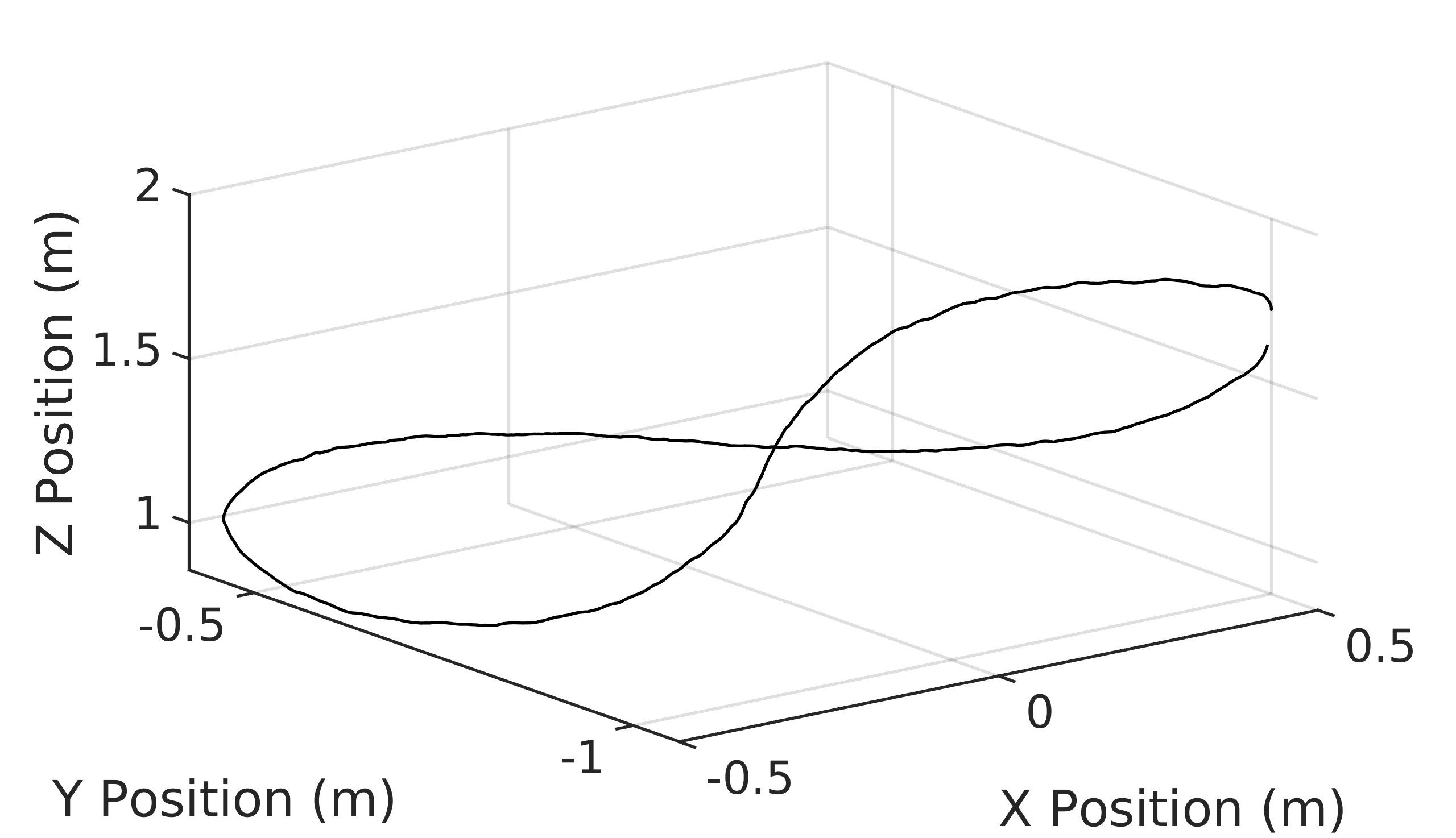}
            \caption{Infinity symbol path-following of thrust vector controller}
            \label{inf}
    \end{figure}

    After the position tracking training, the agent was able to follow any given path without prior training. The path tested was a helical path of height of 1 meter and a 1 meter radius. Fig. \ref{traj2} shows the successful following of the path with a smooth path achieved by the thrust vector controller.
    When compared with the results of the RPM controller, it is clear that the proposed controller achieves a better performance. Fig. \ref{mazinf} shows the response of the RPM controller. A lot of deviations and oscillations are present in the quadrotor's path.
    Another path was passed to ensure the proposed controller's ability to follow any given path. Fig. \ref{inf} shows the proposed controller following an infinity symbol path. The controller followed the path smoothly without any deviations.
     A comprehensive comparison between the proposed thrust vector controller and the RPM controllers proposed in the literature is shown in Table \ref{comp}.

%% file: Conclusion.tex
\section{Conclusion}

This paper introduced a cascaded RL-based control architecture for quadrotors, focusing on controlling the quadrotor's thrust vector instead of the rotors' RPMs directly. Using the SAC algorithm, the proposed controller agent was trained and compared to the RPM controllers used in the literature. Training results showed faster convergence time for the proposed controller. Simulation results showed the better performance of the proposed controller over the conventional RPM controller. The proposed controller stabilizes with 0 steady-state error and follows different paths without deviations. Comparison between the proposed controller and other RPM controllers in the literature was also proposed.